\documentclass[11pt,a4paper]{article}
\usepackage[hyperref]{arr}
\usepackage{times}
\usepackage{footnote}
\makesavenoteenv{tabular}
\usepackage{latexsym}
\hypersetup{
    colorlinks=true, 
    citecolor=blue, 
    urlcolor=black
}
\usepackage[
  separate-uncertainty = true,
  multi-part-units = repeat
]{siunitx}

\usepackage{amsmath, amsthm, amssymb, amsfonts}

\usepackage{url}
\usepackage{arydshln}
\usepackage{times}
\usepackage{array}
\usepackage{multirow,multicol}

\usepackage{graphicx}

\usepackage[english]{babel}

\usepackage{latexsym}

\usepackage{ulem} 

\usepackage{booktabs}
\usepackage{multirow}

\usepackage{enumitem}

\usepackage{graphicx}
\usepackage{subfigure}
\usepackage{stfloats}

\usepackage[hang,flushmargin]{footmisc}

\usepackage[capitalize]{cleveref}

\usepackage{microtype}

\usepackage[T1]{fontenc}

\usepackage[utf8]{inputenc}
\usepackage{booktabs,tabularx}
\usepackage{blindtext}
\usepackage[capitalize]{cleveref}

\newcommand{\model}{BloomNet}
\newcommand{\clo}{course learning outcomes (CLOs)}
\newcommand{\cmmnt}[1]{}
\newenvironment{Table}
  {\par\bigskip\noindent\minipage{\columnwidth}\centering}
  {\endminipage\par\bigskip}

\definecolor{best}{rgb}{0.52, 0.73, 0.4}
\definecolor{better}{rgb}{0.67, 0.88, 0.69}
\definecolor{good}{rgb}{0.64, 0.76, 0.68}
\definecolor{bad}{rgb}{0.93, 0.53, 0.18}
\definecolor{worst}{rgb}{0.89, 0.35, 0.13}

\aclfinalcopy

\title{BloomNet: A Robust Transformer based model for Bloom's Learning Outcome Classification}

\author{Abdul Waheed\(^\alpha\), Muskan Goyal\(^\alpha\), Nimisha Mittal\(^\alpha\), \textbf{Deepak Gupta}\(^\alpha\), \textbf{Ashish Khanna}\(^\alpha\), \\ \textbf{Moolchand Sharma}\(^\alpha\)\\
\(^\alpha\) Maharaja Agrasen Institute of Technology, New Delhi, India. \\
\fontsize{10}{10}\texttt{\{abdulwaheed1513, goyalmuskan1508, nimishamittal1999\}@gmail.com}\\ 
\fontsize{10}{10}\texttt{\{deepakgupta, ashishkhanna, moolchand\}@mait.ac.in}\\ 
}

\date{}

\begin{document}

\maketitle
\begin{abstract}
Bloom's taxonomy is a common paradigm for categorizing educational learning objectives into three learning levels: cognitive, affective, and psychomotor. For the optimization of educational programs, it is crucial to design course learning outcomes (CLOs) according to the different cognitive levels of Bloom's Taxonomy. \cmmnt{The aim is to map the examination questions on the designed CLOs to analyse if the educational program is upto the accreditation standards (stated by NBA).} Usually, administrators of the institutions manually complete the tedious work of mapping CLOs and examination questions to Bloom's taxonomy levels. To address this issue, we propose a transformer based model named \model~that captures linguistic as well semantic information to classify the \clo~. We compare BloomNet with diverse set of basic as well as strong baselines and we observe that our model performs better than all the experimented baselines. Further, we also test the generalisation capability of \model~ by evaluating it on different distributions which our model does not encounter during training and we observe that our model is less susceptible to distribution shift compared to the other considered models. We support our findings by performing extensive result analysis. In ablation study we observe that on explicitly encapsulating the linguistic information along with semantic information improves the model's IID (independent and identically distributed) performance as well as OOD (out-of-distribution) generalization capability. 
\end{abstract}


\section{Introduction}
One of the most difficult challenges faced by the science educators is preparing a curriculum that facilitates the learning process in a structured, planned, and productive manner. It is the goal of the scientific curriculum to educate students who can investigate, question, participate in collaborative projects, and effectively communicate. 
The expected improvements for the students are articulated in a curriculum as learning outcomes \cite{Zorluolu2019AnalyzeOT}. Learning outcomes are used to track, measure, and evaluate the standards and quality of education received by the students at educational institutions \cite{Attia2021BloomsTA}. In terms of these learning outcomes, we may also identify the level of any student. Various measurement and evaluation studies are thus incorporated to determine the level of individual learning outcomes.\par

Exam evaluation is critical for determining how well students understand the course material. Therefore, the objectivity and scientific relevance of the questions developed for exams must be questioned in order to guarantee that students' learning outcomes are tracked and judged effectively. One of the relevant scientific techniques for analyzing this is the Bloom's Taxonomy \cite{Anderson2000ATF}, which is well-known among the educators around the world. The examinations should take account of the difficulty levels, which correspond to the basic objectives and course outcomes in conventional ways like the Bloom’s taxonomy. \par

Dr. Benjamin Bloom, an Educational Psychologist, developed the Bloom’s Taxonomy in 1965. Its goal was to encourage high-order thinking, such as analyzing and examining instead of rote memorization of information \cite{Adesoji2018BloomTO}. The Bloom’s taxonomy is divided into three categories: cognitive (mental skills), affective (emotional areas or attitude), and psychomotor (physical skills). Our study focuses on the cognitive domain, which involves knowledge and intellectual skill development. Researchers have recently demonstrated a growing interest in automatic assessment based on cognitive domains in Bloom's Taxonomy. \cite{Abduljabbar2015ExamQC,Mohammed2018QuestionCB,Yahya2019SwarmIA}. The majority of previous research focused on question classification from a specific domain, while Bloom's taxonomy across the multi-domain region is lacking ways for classifying questions \cite{Sangodiah2017TAXONOMYBF}. This work therefore seeks to establish a question classification method based on the cognitive domain of Bloom’s taxonomy. The Hierarchical order of levels in cognitive domain is: Knowledge, Comprehension, Application, Analysis, Synthesis and Evaluation. The first three levels are categorised as lower level of thinking, whilst the latter three levels are considered as high level of thinking.\par

The primary aim of this study is to assess the utility and efficacy of Bloom's Taxonomy as a framework for establishing course learning outcomes, optimizing curriculum, and evaluating various educational programs. In this paper, we propose BloomNet, a novel transformer-based model that incorporates both linguistic and semantic information for the classification of bloom's course learning outcomes. We also examine the generalisation capability of BloomNet on new distributions because train and test distributions are usually not distributed identically. The evaluation datasets rarely represent the entire distribution and the test distribution often drifts over time \cite{QuioneroCandela2009DatasetSI}, resulting in train-test discrepancies. Due to these discrepancies, models can face unexpected conditions at the test time. Therefore, models should be able to detect and generalise to out-of-distribution (OOD) examples. 

\par In most NLP evaluations, the train and test samples are assumed to be independent and identically distributed (IID). Large pretrained transformer models can achieve high performance on a variety of tasks in the IID scenario \cite{Wang2018GLUEAM}. However, high IID accuracy does not always imply OOD robustness. Furthermore, because pretrained Transformers rely largely on false cues and annotation artifacts \cite{Gururangan2018AnnotationAI, Cai2017PayAT} that OOD instances are less likely to feature, their OOD robustness is unknown. Hence, we examine the robustness of BloomNet and other experimented models such as CNNs, LSTMs, pretrained transformers, and more.


The contributions of our research can be summarized as follows:
\begin{enumerate}
    \item We propose a transformer-based model, BloomNet, that can distinguish between six different cognitive levels of Bloom's taxonomy (Knowledge, Comprehension, Application, Analysis, Synthesis and Evaluation). 
    \item We implement, train and evaluate multiple models to perform comparative analysis.
    \item We evaluate experimented models for OOD generalization and we observe that pretrained transformers  (RoBERTa, DistilRoBERTa \cite{Liu2019RoBERTaAR, Sanh2019DistilBERTAD}) along with proposed model have better generalization capability compared to other models.
    \item We perform ablation study to asses the contribution of various components in proposed model.
\end{enumerate}

The following is the final exhibition. Section 2 and Section 3 describes the previous work and methodology respectively. Section 4 delves into the experiments and results. The conclusion and possible future directions are discussed in Section 5.

\subsection*{Reproducibility} We implement, train and evaluate all the considered models as well as \model~ as described in \ref{setup}. We intend to release all the code, trained weights, datasets,  and configuration upon acceptance.

\section{Related Work}
Text classification is an important NLP research area with numerous applications. A number of scholars have concentrated on automatic text classification. In recent years, classification of exam questions for the cognitive domain of Bloom's taxonomy has received a lot of attention. Previous works have used different features and methods for text classification. Some of these works are discussed in this section. 

In \cite{Chang2009AutomaticAB}, an online examination system is created that supports automatic Bloom's taxonomy analysis for the test questions. The researchers introduce fourteen keywords for the analysis on questions. Each keyword is associated with a specific cognition level. The experiment is conducted on 288 test items and a 75\% accuracy is achieved for the "Knowledge" cognition level. \par 

A. Swart and M. Daneti \cite{Swart2019AnalyzingLO} analyzed the learning outcomes for Electronic fundamental module (of two universities) using Bloom's Taxonomy. To identify the proportion of each cognition level, the verbs of each learning outcome are connected to certain specific verbs in Bloom's taxonomy. This reflected the balance between theory and practice for the cognitive development of electrical engineering students. The consistency of the findings of the two universities demonstrated that students could blend theory and practice because they had around 40 percent of higher level cognitive outcomes.\par 

Likewise, \cite{Mohammed2020QuestionCB} classified exam questions for the cognitive domain of Bloom's Taxonomy using TFPOS-IDF and pre-trained word2vec. To classify the questions, the extracted features are fed to three distinct classifiers i.e. logistic regression, K-nearest neighbour, and Support Vector Machine. For the experiment, they employ two datasets, one with 141 questions and the other with 600 questions. The first dataset results in a weighted average of 71.1\%, 82.3\% and 83.7\% while the second achieves a weighted average of 85.4\%, 89.4\% and 89.7\%. \par 

Adidah Lajis et al. proposed \cite{Lajis2018ProposedAF} a framework for assessing students' programming skills. Bloom's taxonomy cognitive domain serves as the foundation for the framework. According to the findings, Bloom's taxonomy could be used as a basis for grading students. It said that the students would be judged based on their ability using Bloom's taxonomy. The authors also suggested that taxonomy be used as an evaluation framework rather than learning. \par 

Based on their domain knowledge, teachers and accreditation organizations manually classify course learning outcomes (CLOs) and questions on distinct levels of cognitive domain. This is time-consuming and usually leads in errors due to human bias. As a result, this technique must be automated. Several scholars have sought to automate this process through the use of natural language processing and machine learning techniques \cite{Haris2012ARA,Jayakodi2015AnAC, Osadi2017EnsembleCB, Kowsari2019TextCA}. Deep learning has recently exhibited impressive results when compared to traditional machine learning methods, particularly in the field of text classification \cite{Minaee2020DeepLT}. \par

For text classification tasks, several neural models that automatically represent text as embedding have been developed, such as CNNs, RNNs, graph neural networks, and a variety of attention models such as hierarchical attention networks, self-attention networks, and so on. The majority of previous efforts on Bloom's taxonomy have either used traditional machine learning approaches or representative deep neural models such as RNNs, LSTMs, and so on. In this research, we propose a transformer-based approach for performing text classification as per cognitive domains. Transformers, \cite{Vaswani2017AttentionIA} provide significantly better parallelization than RNNs, allowing for efficient (pre-)training of very large language models and an enhanced performance rate.\par

\begin{figure*}[h!]
    \centering
    \includegraphics[scale=0.60]{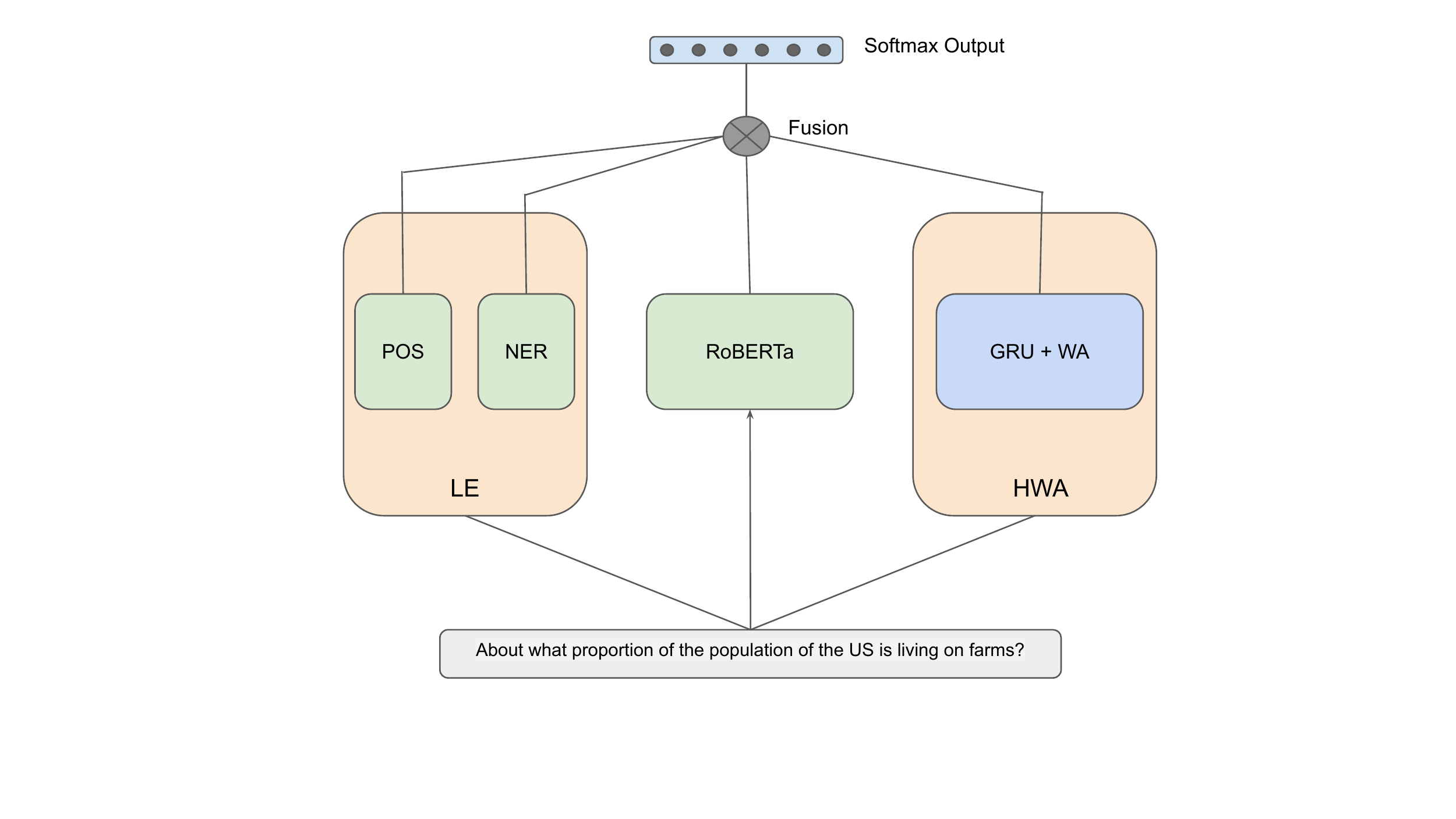}
    \caption{A high level architectrue digram of the proposed model \model~. POS (Part of Speech} Module. NER(Named Entity Recognition) Module. HWA (Hierarchical Word Attention) Module. LE (Linguistic Encapsulation).
    \label{fig:my_label}
\end{figure*}

\section{Methodology}
In this section we discuss the methodology part of our research. Our model is inspired by \cite{gupta-etal-2021-lesa} and \cite{yang-etal-2016-hierarchical}. In \model~, we encapsulate the linguistic information along with generic input representation and we also explicitly model word level attention. In following sections we describe each component of our model (shown in Figure \ref{fig:my_label}) in detail.\\
\textbf{Notation:}  We denote current input as set of tokens $ x \in X = \{t_0, t_1, t_2, ...t_n\}$ where $n$ is the number of tokens in input. We define a model as $f_\texttt{model}:x \longrightarrow h$  where $h \in \mathbb{R}^{d}$. We define our final classfieir as $f_\texttt{c}:x \longrightarrow C$  where $C$ is the softmax output and its size is equal to number of classes in our data.

\subsection{Representation Model}
Representation model or language encoder is main component of \model~ which gives contextualized embeddings \cite{devlin-etal-2019-bert, Pennington2014GloVeGV} for the text input. and for this we use pretrained RoBERTa \cite{Liu2019RoBERTaAR} model from huggingface model hub repository \cite{wolf-etal-2020-transformers}. The reason we use RoBERTa instead of its other widely used counterparts such as BERT \cite{devlin-etal-2019-bert}, DistilBERT \cite{Sanh2019DistilBERTAD} is that it has seen much more data during its pretraining compare to its predecessor which results in increased robustness for subpopulation as well distribution shift. We feed tokenized input to RoBERTa model and we use \texttt{CLS} token as input representation. It can be represented as :

\begin{equation}
    h_\text{rep} = f_\text{rep}(x)  \hfill
\end{equation}


\subsection{Linguistic Encapsulation}
Work by \cite{gupta-etal-2021-lesa} shows that explicit encapsulation of linguistic information increases the performance of the model for claim detection task, inspired by the same we also explicitly encapsulate linguistic information in modelling of \model. We use POS (Part-Of-Speech) and NER (Named Entity Recognition) information coming from a trained model for POS and NER tasks respectively. We freeze the POS and NER model during training so that it's weights do not change and hence it carries linguistic information. We use \texttt{CLS} token representation of the model and we define it as as follows:

\begin{equation}
    h_\text{POS} = f_\text{POS}(x) 
\end{equation}

\begin{equation}
    h_\text{NER} = f_\text{NER}(x) 
\end{equation}

\subsection{Hierarchical Word Attention and Classification}
Inspired by \cite{yang-etal-2016-hierarchical} we use word level attention to get the better dense representation of the input. For this we use GRU \citet{cho-etal-2014-learning} and apply word level attention on its output. As result we get a vector from this module as input representation and we use this along with other information for classification. We denote this as follows :
\begin{equation}
    h_\text{HWA} = f_\text{HWA}(x) 
\end{equation}

Finally, we get four different representation coming from different components and we fuse these information using concatenation and feed this to a linear classification model. We write the concatenation as:
\begin{equation}
    H = h_\text{Rep} \oplus h_\text{POS} \oplus h_\text{NER} \oplus h_\text{HWA}
\end{equation}
Classification module can be represented as:
\begin{equation}
    C = f_\text{c}(H)
\end{equation}


\section{Experiments and Results}

\subsection{Dataset}
We use two open domain datasets to evaluate the proposed approach. First dataset was proposed in \cite{Yahya2012BloomsTC} which comprises 600 open-ended questions. The second dataset was compiled from a variety of websites, publications, and previous research \cite{Haris2015BloomsTQ}. It contains 141 open-ended questions. The datasets are annotated and classified into six categories (Knowledge, Comprehension, Application, Analysis, Synthesis, and Evaluation). Table \ref{tab:dataset} illustrates the label distribution for both datasets. The questions in these two datasets come from a variety of fields of study, including chemical, literature, biological, artistic, and computer science, among others. 

\begin{table}[h]
\def\arraystretch{1.5}%
\begin{Table}
\begin{tabular} [l]{ c c c } 
    \toprule[1.5pt]
     \textbf{Cognitive Level} &  \textbf{Dataset 1} & \textbf{Dataset 2}\\
     \hline
     Knowledge Level & 100 & 26\\
     Comprehension Level & 100 & 23\\
     Application Level & 100 & 15\\
     Analysis Level & 100 & 23\\
     Synthesis Level & 100 & 30\\
     Evaluation Level & 100 & 24\\
     \hline \textbf{Total} & \textbf{600} & \textbf{141}\\
    \bottomrule[1.5pt]
\end{tabular}
\caption{Number of questions in each cognitive level}
\label{tab:dataset}
\end{Table}
\end{table}

\subsection{Baselines}
In this section, we describe the various baseline models that we used for comparative analysis. These models are arranged in the order of their performance.

\subsubsection{VDCNN}
\par Very Deep CNN (VDCNN) \cite{Schwenk2017VeryDC} learns a hierarchical representation of a sentence with the help of a deep stack of convolutions and max-pooling of size 3 and by operating at the character level representation of the text. VDCNNs are substantially deeper than convolutional neural networks published previously. This is the first CNN model to present the "advantage of depth" in the field of NLP.

\subsubsection{LSTM}
\par Text is viewed as a sequence of words in RNN-based models, which are designed to capture word dependencies and text structures for text classification. RNNs \cite{10.5555/553011} can memorise the local structure of a word sequence, but they struggle with long-range dependencies. Long-Short Term Memory (LSTM) \cite{Sari2020TextCU} is the most popular variant of RNN, that is created to capture long term dependencies. Vanilla RNNs suffer from gradient vanishing problem and LSTMs resolve this issue by using a memory cell that remember values across arbitrary time periods.

\subsubsection{HAN}
\par Hierarchical Attention Networks (HAN) \cite{Yang2016HierarchicalAN} collects  relevant tokens from sentences and aggregate their representation with the help of an attention mechanism. The same approach is used to retrieve relevant sentence vectors that is used in the classification task. 

\subsubsection{CNN}
\par RNNs are taught to detect patterns over time, while CNNs \cite{Kim2014ConvolutionalNN}  are taught to recognise patterns over space. RNNs work for NLP tasks like POS tagging or QA that need understanding of long-range semantics, but CNNs are good for recognising local and position-invariant patterns \cite{LeCun1998GradientbasedLA}. These patterns could be key phrases expressing a specific emotion or a topic. As a result, CNNs have become one of the most common text classification model.

\subsubsection{RCNN} 
\par In contrast to CNN, Recurrent CNN \cite{Girshick2014RichFH} comprises of bi-directional recurrent structure that captures greater contextual data from word representations. This is followed by a max pooling layer which is responsible for extracting key features for text classification.

\subsubsection{Seq2Seq-Attention} 
\par Deep learning models known as sequence-to-sequence \cite{Bahdanau2015NeuralMT} models have been deployed in tasks such as machine translation, text summarization, and image captioning. Seq2Seq comprises of encoder, decoder and attention layer where encoder is responsible for compiling data in the form of vector. Further this context is parsed to the decoder that produces desired output sequence.
The primary idea behind the attention mechanism is to avoid learning a single vector representation for each sentence and instead be attentive to specific input vectors based on the attention weights.

\subsubsection{Self-Attention}
Self-attention is a type of attention that allows us to learn the relationship between words in a sentence. Various NLP tasks and Transformers \cite{Vaswani2017AttentionIA} use self-attention. Despite the fact that CNNs are less sequential than RNNs, the computing cost of capturing relationships between words in a phrase increases with the length of the sentence, much like RNNs. Transformers get around this constraint by using self-attention to compute a "attention score" for each word in a sentence or document in parallel, modelling the influence each word has on the others.

\subsubsection{TF-IDF Random Forest} 
\par Random Forest (RF) models \cite{Xue2015ResearchOT} are made up of a collection of decision trees that were trained on random feature subsets. This model's predictions are obtained via a majority vote of all forest tree projections. In addition, RF classifiers are simple to apply to text classification of high-dimensional noisy data. Furthermore, TF-IDF (Term Frequency Inverse Document Frequency) \cite{ref1} is a commonly used approach for converting text to a number representation that may be employed by a machine algorithm. TfidfVectorizer weights word counts based on how frequently they appear in the sentence.

\subsubsection{DistilRoBERTa} 
DistilRoBERTa has been distilled from RoBERTa-base model \cite{Liu2019RoBERTaAR,Sanh2019DistilBERTAD} that contains around half number of parameters as BERT model. It is based on same training process as that of DistilBERT. Moreover, DistilRoBERTa maintains 95 percent of BERT's performance on the GLUE language understanding benchmark \cite{Wang2018GLUEAM}.

\subsubsection{RoBERTa} 
\par RoBERTa (Robustly Optimized BERT Pretraining Accuracy) model \cite{Liu2019RoBERTaAR} is a more robust version of BERT that is trained with a lot more data. It is based on fine-tuning the hyper-parameters that has improved the results and performance of the model significantly. To boost performance of BERT, RoBERTa also modified its training procedure and architecture. These modifications include removing next sentence prediction and dynamically changing the masking pattern during pre-training.

\subsection{Experimental Setup}
\label{setup}
We use HuggingFace Transformers \cite{wolf-etal-2020-transformers}, PyTorch \cite{NEURIPS2019_bdbca288}, PyTorch-Lightning \cite{falcon2019pytorch} and Scikit-learn \cite{JMLR:v12:pedregosa11a} for model implementation, training and evaluation. We train all of the models with KFold (k=5) cross validation and report \texttt{mean} and \texttt{std} values across the folds. We observe that the text in the dataset is relatively short, thus we use \texttt{maximum sequence lengths = 128}. For LSTM-like models, we employ \texttt{hidden size = 768}, \texttt{number of layers = 4}, and \texttt{dropout = 0.10} throughout the experiments. We use \texttt{Adam} \cite{Kingma2015AdamAM} as the optimizer and \texttt{cross-entropy} as the objective function. We use different \texttt{learning rates} for different models depending on how they are initialized, and for \texttt{\model} we use \texttt{learning rate = 2e-5} and train all models for \texttt{50 epochs} with \texttt{batch size = 32} and \texttt{early stopping} to prevent overfitting. We do not change the shared hyper-parameters across the models so that the comparison is as fair as possible. We do not conduct comprehensive hyper-parameter searches due to computational constraints.

\subsection{Results}
We evaluate following baselines to compare the performance of our proposed model, BloomNet: VDCNN \cite{Schwenk2017VeryDC}, LSTM \cite{Sari2020TextCU}, HAN\cite{Yang2016HierarchicalAN}, CNN\cite{Kim2014ConvolutionalNN}, RCNN \cite{Girshick2014RichFH}, Seq2Seq-Attention \cite{Bahdanau2015NeuralMT}, Self-Attention \cite{Vaswani2017AttentionIA}, Random Forest \cite{Xue2015ResearchOT}, DistilRoBERTa \cite{Liu2019RoBERTaAR}, and RoBERTa \cite{Liu2019RoBERTaAR}. We find that BloomNet outperforms all the considered baselines on the two datasets, demonstrating its higher performance for text classification. Table \ref{tab:results} reports the performance of BloomNet and all the baselines.  
\par The model is trained on Dataset1 and evaluated for both the datasets. Dataset1 is used for evaluating BloomNet's IID performance while Dataset2 is used to test BloomNet's generalisation capabilities (OOD performance) by assessing it on new distributions that our model does not encounter during the training process. We observe that in comparison to the baseline models, BloomNet is less vulnerable to distribution shift.

\begin{savenotes}
    \begin{table*}[h]
        \centering
        \begin{tabular}[l]{c c c c c}
        \toprule[1.5pt]
            \multirow{2}{*}{\textbf{Model}} &
            \multicolumn{2}{c}{\textbf{Dataset1 (IID)}} & 
            \multicolumn{2}{c}{\textbf{Dataset2(OOD)}} \\
            & \bf Accuracy & \bf Macro-F1 & \bf Accuracy & \bf Macro-F1 \\
            \toprule[1 pt]
             
            \text{VDCNN\footnote{Very Deep Convolutional Networks for Text Classification(VDCNN)}} & ${ 32.00\pm6.78 }$ & ${ 31.70\pm6.71 }$ & ${ 28.79\pm3.82 }$ & ${26.54 \pm4.12 }$  \\
            
             \text{LSTM\footnote{Long Short-Term Memory (LSTM)}} & ${58.50 \pm 3.99 }$  & ${ 59.27 \pm 3.55 }$ & ${ 47.09 \pm 4.05 }$ & ${ 45.47\pm2.71 }$ \\
             

            \text{HAN\footnote{Hierarchical Attention Networks (HAN)}} & ${ 59.64\pm 3.72}$ & ${ 58.90\pm 4.16}$  & ${ 54.69 \pm3.39 }$  & ${50.61 \pm 3.12}$  \\
            
             \text{CNN\footnote{Convolutional Neural Network (CNN)}} & ${ 60.67\pm1.11 }$  & ${ 60.57\pm1.36 }$ & ${ 49.79\pm2.17 }$ & ${48.17 \pm 2.00 }$ \\
             
             \text{RCNN\footnote{Recurrent Convolutional Neural Network (RCNN)}} & ${ 66.33\pm3.01 }$ & ${ 65.90\pm3.51 }$ & ${ 54.04\pm 3.57}$ & ${ 51.05\pm3.09 }$  \\
             
            \text{Seq2Seq-Attention\footnote{Sequential to Sequential Model with Attention}} & ${ 64.00 \pm 3.09}$ & ${ 63.79 \pm 3.50}$ & ${ 52.91 \pm 2.22}$ & ${ 50.92 \pm 2.11 }$  \\
            
             \text{Self-Attention} & ${70.17 \pm 3.55 }$ & ${ 69.92 \pm 3.80 }$ & ${ 55.46 \pm 2.07 }$ & ${ 52.75 \pm 1.81 }$  \\
             
             \text{Random Forest TF-IDF\footnote{Term Frequency - Inverse Document Frequency(TF-IDF)}} & ${ 70.66\pm2.52 }$ & ${ 70.50\pm 2.75}$  & ${ 62.12\pm1.38 }$  & ${ 58.04\pm1.73 }$  \\
             
             
             
            \text{DistilRoBERTa\footnote{Distilled from RoBERTa model}} & ${ 80.50\pm3.23 }$ & ${ 80.21\pm3.49 }$ & ${ 67.80\pm 1.59}$ & ${ 63.94\pm1.48 }$ \\
             
             \text{RoBERTa\footnote{Robustly Optimized BERT Pre-training Approach}} & ${82.00 \pm 2.01}$ & ${81.67 \pm 2.20 }$  & ${68.65 \pm 2.74 }$  & ${65.65\pm 2.82}$  \\
             
             \midrule[0.2 pt]
               
             
             \text{BloomNet} & ${ \textbf{87.50} \pm \textbf{1.88} }$ & ${ \textbf{87.23} \pm \textbf{2.47} }$  & ${ \textbf{70.40}\pm\textbf{2.52}}$  & ${ \textbf{67.10}\pm \textbf{2.43}}$  \\
             
        \bottomrule[1.5pt]
        \end{tabular}
        \caption{Mean and Standard deviation of the results obtained over 5 folds. \model~performs significantly better (p < 0.004) than the RoBERTa. \textbf{Bold} shows best performance. All models are trained and evaluation on Dataset1 hence IID, and OOD evaluation is performed on Dataset2. }
        \label{tab:results}
    \end{table*}
\end{savenotes}

\subsubsection{Comparative Analysis}
As seen in Table \ref{tab:results} BloomNet outperforms the baseline models and achieves ${87.50 \pm 1.88}$ and ${70.40 \pm 2.52}$ and Macro-F1 score ${87.23 \pm 2.47}$ and ${67.10 \pm 2.43}$ on Dataset1(IID) and Dataset2(OOD) respectively. 

\par In addition, we also made some very in-depth observations while evaluating the baselines. Surprisingly, the TF-IDF \cite{ref1} encoded text with random forest performs better than several strong baselines like LSTM, HAN, CNN, and RCNN. It is the third best performing baseline model that achieves ${70.66 \pm 2.52}$ and ${62.12 \pm 1.38}$ accuracy and Macro-F1 ${70.50 \pm 2.75}$ and ${58.04 \pm 1.73}$ on Dataset1(IID) and Dataset2(OOD) respectively. \par
We also observe that Attention based models like Seq2Seq-Attention and Self-Attention show better classification performance than vanilla models (like VDCNN, CNN, LSTM, and RCNN). Further, we investigate BERT-based models DistilRoBERTa and RoBERTa (which are pretrained Transformers) that achieve superior performanc over all the other considered baselines. RoBERTa is the best performing model with accuracy of ${82.00 \pm 2.01}$ and ${68.65 \pm 2.74}$ and Macro-F1 score ${81.67 \pm 2.20}$ and ${65.65 \pm 2.82}$ on Dataset1(IID) and Dataset2(OOD) respectively. 

\subsubsection{Out-of-distribution Generalisation}
We evaluate models on new data which is not seen during training to evaluate the OOD robustness. We observe that OOD and IID performance is linearly correlated. The models that do not perform well on IID data such as $\text{VDCNN, LSTM, etc}$ also perform poor on OOD data. Pretrained transformers have been proven robust to distribution shift \cite{hendrycks-etal-2020-pretrained, ramesh-kashyap-etal-2021-analyzing} but in our case we notice significant performance drop ($~20\%$) between IID and OOD data across all the pretrained transformer based models in our experiment which is same for other models as well. We hypothesise that this might be caused by large discrepancy between IID and OOD data.  

\subsection{Ablation Study}
\begin{table}[]
    \centering
    \begin{tabular}{c  c c}
        \toprule[1.5pt]
         \textbf{Component} & \textbf{Accuracy} & \textbf{Macro-F1} \\
         \midrule[1pt]
         RoBERTa & 82.00 & 81.67 \\
         + WA & 84.11  & 84.10 \\
         + POS-NER & 84.64 & 84.48 \\
         + WA + POS-NER & \textbf{87.50} & \textbf{87.23} \\
        \bottomrule[1.5pt]
    \end{tabular}
    \caption{Ablation results for \model~. Mean of IID Accuracy and Macro-F1 is reported. Linguistic Encapsulation along with Word level attention yields significantly better (p <0.004) results. WA: Word Attention. POS: Part-of-speech. NER: Named-Entity Recognition. }
    \label{tab:ablation}
\end{table}

Our proposed model \model~ has three main component: 1. Representation model or Language Encoder 2. Linguistic Encapsulation Module and 3. Hierarchical Word Attention Model. We conduct an ablation study to assess the contribution of different components in our model.
First we remove the word attention module from \model~ and train it like other models with same configuration. We observe the \model~ without word attention yields $\approx 84$ and $\approx 65$ accuracy for IID and OOD data respectively. Then we remove the linguistic encapsulation block and train the model like previously. \model~ without linguistic encapsulation yeilds similar IID performance ($\approx 84$ accuracy) but performs better on OOD data. If we remove the both components word attention as well as linguistic encapsulation \model~ is same as RoBERTa \cite{Liu2019RoBERTaAR} baseline. The result of ablation is stated in the table \ref{tab:ablation}. 


\section{Discussion}
\textbf{Limitations:} We propose a novel transformer based model named \model~ which has three language encoder (we use RoBERTa), two for linguistic encoding named as POS Encoder and NER Encoder, and one generic encoder. Due to three large transformer based language encoder proposed model is compute and memory heavy hence it becomes very cumbersome to deploy it in production. To asses the generalization capability of models we evaluate them on a different distribution which they do not see during training. We do not quantify the shift between IID and OOD and we restrict ourselves to only evaluation as investing the cause of performance drop on OOD data is beyond scope of this study. The datasets used in our work is relatively small having 600 and 141 samples respectively in both Dataset1 and Dataset2. Although we do cross validation and report mean and standard-deviation but we expect change in performance on bigger dataset. For same reason we do not train models on Dataset2.  \\
\textbf{\\Ethical Considerations:} We are well aware of the societal implication of deploying large language models it could have unintended bias against marginalized groups and model itself plays significant role in amplifying those biases. We do not see any immediate misuse of our work, but more research in this area could lead to the development of systems such as automated scoring, which can have a disproportionately detrimental impact on marginalized groups.

\section{Conclusion and Future Work} 
We propose a novel transformer-based model, BloomNet, that captures the linguistic and semantic information to classify the course learning outcomes according to the different cognitive domains of Bloom's Taxonomy. BloomNet outperforms the considered baseline models analyzed in this study in terms of performance and generalization capability. Interestingly, we observe that carefully processed text with TF-IDF encoding outperforms numerous strong baselines like CNN, RNN, and attention based models. We also observe that pretrained Transformers generalize to OOD examples surprisingly well. Overall, we use a state-of-the-art Natural Language Processing (NLP) model for a relatively new task, and we believe it opens up new research directions for NLP in the education domain. We believe that, similar to previous domain-oriented NLP studies, such as NLP4Health, NLP4Programming, LegalNLP, and so on, NLP4Education has the potential to improve existing systems for the mutual benefit of the community and society in general. This is a novel task employing the state-of-the-art Natural Language Processing(NLP) system into education which is relatively new and we believe that it will open a new direction for NLP research.



\bibliographystyle{acl_natbib}
\bibliography{arr}

\begin{thebibliography}{48}
\expandafter\ifx\csname natexlab\endcsname\relax\def\natexlab#1{#1}\fi

\bibitem[{Abduljabbar and Omar(2015)}]{Abduljabbar2015ExamQC}
D.~Abduljabbar and N.~Omar. 2015.
\newblock Exam questions classification based on bloom’s taxonomy cognitive
  level using classifiers combination.
\newblock \emph{Journal of theoretical and applied information technology},
  78:447--455.

\bibitem[{Adesoji(2018)}]{Adesoji2018BloomTO}
F.~Adesoji. 2018.
\newblock Bloom taxonomy of educational objectives and the modification of
  cognitive levels.
\newblock \emph{Advances in Social Sciences Research Journal}, 5.

\bibitem[{Anderson et~al.(2000)Anderson, Krathwohl, and
  Bloom}]{Anderson2000ATF}
L.~Anderson, D.~Krathwohl, and B.~Bloom. 2000.
\newblock A taxonomy for learning, teaching, and assessing: A revision of
  bloom's taxonomy of educational objectives.

\bibitem[{Attia(2021)}]{Attia2021BloomsTA}
A.~S. Attia. 2021.
\newblock Bloom’s taxonomy as a tool to optimize course learning outcomes and
  assessments in architecture programs.

\bibitem[{Bahdanau et~al.(2015)Bahdanau, Cho, and
  Bengio}]{Bahdanau2015NeuralMT}
Dzmitry Bahdanau, Kyunghyun Cho, and Yoshua Bengio. 2015.
\newblock Neural machine translation by jointly learning to align and
  translate.
\newblock \emph{CoRR}, abs/1409.0473.

\bibitem[{Cai et~al.(2017)Cai, Tu, and Gimpel}]{Cai2017PayAT}
Zheng Cai, Lifu Tu, and Kevin Gimpel. 2017.
\newblock Pay attention to the ending: Strong neural baselines for the roc
  story cloze task.
\newblock In \emph{ACL}.

\bibitem[{Chang and Chung(2009)}]{Chang2009AutomaticAB}
Wen-Chih Chang and Ming-Shun Chung. 2009.
\newblock Automatic applying bloom's taxonomy to classify and analysis the
  cognition level of english question items.
\newblock \emph{2009 Joint Conferences on Pervasive Computing (JCPC)}, pages
  727--734.

\bibitem[{Cho et~al.(2014)Cho, van Merri{\"e}nboer, Gulcehre, Bahdanau,
  Bougares, Schwenk, and Bengio}]{cho-etal-2014-learning}
Kyunghyun Cho, Bart van Merri{\"e}nboer, Caglar Gulcehre, Dzmitry Bahdanau,
  Fethi Bougares, Holger Schwenk, and Yoshua Bengio. 2014.
\newblock \href {https://doi.org/10.3115/v1/D14-1179} {Learning phrase
  representations using {RNN} encoder{--}decoder for statistical machine
  translation}.
\newblock In \emph{Proceedings of the 2014 Conference on Empirical Methods in
  Natural Language Processing ({EMNLP})}, pages 1724--1734, Doha, Qatar.
  Association for Computational Linguistics.

\bibitem[{Devlin et~al.(2019)Devlin, Chang, Lee, and
  Toutanova}]{devlin-etal-2019-bert}
Jacob Devlin, Ming-Wei Chang, Kenton Lee, and Kristina Toutanova. 2019.
\newblock \href {https://doi.org/10.18653/v1/N19-1423} {{BERT}: Pre-training of
  deep bidirectional transformers for language understanding}.
\newblock In \emph{Proceedings of the 2019 Conference of the North {A}merican
  Chapter of the Association for Computational Linguistics: Human Language
  Technologies, Volume 1 (Long and Short Papers)}, pages 4171--4186,
  Minneapolis, Minnesota. Association for Computational Linguistics.

\bibitem[{Falcon(2019)}]{falcon2019pytorch}
et~al. Falcon, WA. 2019.
\newblock Pytorch lightning.
\newblock \emph{GitHub. Note:
  https://github.com/PyTorchLightning/pytorch-lightning}, 3.

\bibitem[{Girshick et~al.(2014)Girshick, Donahue, Darrell, and
  Malik}]{Girshick2014RichFH}
Ross~B. Girshick, Jeff Donahue, Trevor Darrell, and J.~Malik. 2014.
\newblock Rich feature hierarchies for accurate object detection and semantic
  segmentation.
\newblock \emph{2014 IEEE Conference on Computer Vision and Pattern
  Recognition}, pages 580--587.

\bibitem[{Gupta et~al.(2021)Gupta, Singh, Sundriyal, Akhtar, and
  Chakraborty}]{gupta-etal-2021-lesa}
Shreya Gupta, Parantak Singh, Megha Sundriyal, Md.~Shad Akhtar, and Tanmoy
  Chakraborty. 2021.
\newblock \href {https://aclanthology.org/2021.eacl-main.277} {{LESA}:
  Linguistic encapsulation and semantic amalgamation based generalised claim
  detection from online content}.
\newblock In \emph{Proceedings of the 16th Conference of the European Chapter
  of the Association for Computational Linguistics: Main Volume}, pages
  3178--3188, Online. Association for Computational Linguistics.

\bibitem[{Gururangan et~al.(2018)Gururangan, Swayamdipta, Levy, Schwartz,
  Bowman, and Smith}]{Gururangan2018AnnotationAI}
Suchin Gururangan, Swabha Swayamdipta, Omer Levy, Roy Schwartz, Samuel~R.
  Bowman, and Noah~A. Smith. 2018.
\newblock Annotation artifacts in natural language inference data.
\newblock In \emph{NAACL-HLT}.

\bibitem[{Haris and Omar(2012)}]{Haris2012ARA}
S.~S. Haris and N.~Omar. 2012.
\newblock A rule-based approach in bloom's taxonomy question classification
  through natural language processing.
\newblock \emph{2012 7th International Conference on Computing and Convergence
  Technology (ICCCT)}, pages 410--414.

\bibitem[{Haris and Omar(2015)}]{Haris2015BloomsTQ}
S.~S. Haris and N.~Omar. 2015.
\newblock Bloom's taxonomy question categorization using rules and n-gram
  approach.
\newblock \emph{Journal of theoretical and applied information technology},
  76:401--407.

\bibitem[{Hendrycks et~al.(2020)Hendrycks, Liu, Wallace, Dziedzic, Krishnan,
  and Song}]{hendrycks-etal-2020-pretrained}
Dan Hendrycks, Xiaoyuan Liu, Eric Wallace, Adam Dziedzic, Rishabh Krishnan, and
  Dawn Song. 2020.
\newblock \href {https://doi.org/10.18653/v1/2020.acl-main.244} {Pretrained
  transformers improve out-of-distribution robustness}.
\newblock In \emph{Proceedings of the 58th Annual Meeting of the Association
  for Computational Linguistics}, pages 2744--2751, Online. Association for
  Computational Linguistics.

\bibitem[{Jain and Medsker(1999)}]{10.5555/553011}
L.~C. Jain and L.~R. Medsker. 1999.
\newblock \emph{Recurrent Neural Networks: Design and Applications}, 1st
  edition.
\newblock CRC Press, Inc., USA.

\bibitem[{Jayakodi et~al.(2015)Jayakodi, Bandara, and
  Perera}]{Jayakodi2015AnAC}
K.~Jayakodi, M.~Bandara, and I.~Perera. 2015.
\newblock An automatic classifier for exam questions in engineering: A process
  for bloom's taxonomy.
\newblock \emph{2015 IEEE International Conference on Teaching, Assessment, and
  Learning for Engineering (TALE)}, pages 195--202.

\bibitem[{Kim(2014)}]{Kim2014ConvolutionalNN}
Yoon Kim. 2014.
\newblock Convolutional neural networks for sentence classification.
\newblock In \emph{EMNLP}.

\bibitem[{Kingma and Ba(2015)}]{Kingma2015AdamAM}
Diederik~P. Kingma and Jimmy Ba. 2015.
\newblock Adam: A method for stochastic optimization.
\newblock \emph{CoRR}, abs/1412.6980.

\bibitem[{Kowsari et~al.(2019)Kowsari, Meimandi, Heidarysafa, Mendu, Barnes,
  and Brown}]{Kowsari2019TextCA}
Kamran Kowsari, K.~Meimandi, Mojtaba Heidarysafa, Sanjana Mendu, Laura~E.
  Barnes, and D.~Brown. 2019.
\newblock Text classification algorithms: A survey.
\newblock \emph{Inf.}, 10:150.

\bibitem[{Lajis et~al.(2018)Lajis, Nasir, and Aziz}]{Lajis2018ProposedAF}
Adidah Lajis, H.~Nasir, and N.~A. Aziz. 2018.
\newblock Proposed assessment framework based on bloom taxonomy cognitive
  competency: Introduction to programming.
\newblock \emph{Proceedings of the 2018 7th International Conference on
  Software and Computer Applications}.

\bibitem[{LeCun et~al.(1998)LeCun, Bottou, Bengio, and
  Haffner}]{LeCun1998GradientbasedLA}
Y.~LeCun, L.~Bottou, Yoshua Bengio, and P.~Haffner. 1998.
\newblock Gradient-based learning applied to document recognition.

\bibitem[{Liu et~al.(2019)Liu, Ott, Goyal, Du, Joshi, Chen, Levy, Lewis,
  Zettlemoyer, and Stoyanov}]{Liu2019RoBERTaAR}
Yinhan Liu, Myle Ott, Naman Goyal, Jingfei Du, Mandar Joshi, Danqi Chen, Omer
  Levy, M.~Lewis, Luke Zettlemoyer, and Veselin Stoyanov. 2019.
\newblock Roberta: A robustly optimized bert pretraining approach.
\newblock \emph{ArXiv}, abs/1907.11692.

\bibitem[{Minaee et~al.(2020)Minaee, Kalchbrenner, Cambria, Nikzad, Chenaghlu,
  and Gao}]{Minaee2020DeepLT}
Shervin Minaee, Nal Kalchbrenner, E.~Cambria, Narjes Nikzad, M.~Chenaghlu, and
  Jianfeng Gao. 2020.
\newblock Deep learning--based text classification.
\newblock \emph{ACM Computing Surveys (CSUR)}, 54:1 -- 40.

\bibitem[{Mohammed and Omar(2018)}]{Mohammed2018QuestionCB}
Manal Mohammed and N.~Omar. 2018.
\newblock Question classification based on bloom’s taxonomy using enhanced
  tf-idf.
\newblock \emph{International Journal on Advanced Science, Engineering and
  Information Technology}, 8:1679--1685.

\bibitem[{Mohammed and Omar(2020)}]{Mohammed2020QuestionCB}
Manal Mohammed and N.~Omar. 2020.
\newblock Question classification based on bloom’s taxonomy cognitive domain
  using modified tf-idf and word2vec.
\newblock \emph{PLoS ONE}, 15.

\bibitem[{Osadi et~al.(2017)Osadi, Fernando, and Welgama}]{Osadi2017EnsembleCB}
K.~A. Osadi, Mgnas Fernando, and W.~V. Welgama. 2017.
\newblock Ensemble classifier based approach for classification of examination
  questions into bloom’s taxonomy cognitive levels.
\newblock \emph{International Journal of Computer Applications}, 162:1--6.

\bibitem[{Paszke et~al.(2019)Paszke, Gross, Massa, Lerer, Bradbury, Chanan,
  Killeen, Lin, Gimelshein, Antiga, Desmaison, Kopf, Yang, DeVito, Raison,
  Tejani, Chilamkurthy, Steiner, Fang, Bai, and
  Chintala}]{NEURIPS2019_bdbca288}
Adam Paszke, Sam Gross, Francisco Massa, Adam Lerer, James Bradbury, Gregory
  Chanan, Trevor Killeen, Zeming Lin, Natalia Gimelshein, Luca Antiga, Alban
  Desmaison, Andreas Kopf, Edward Yang, Zachary DeVito, Martin Raison, Alykhan
  Tejani, Sasank Chilamkurthy, Benoit Steiner, Lu~Fang, Junjie Bai, and Soumith
  Chintala. 2019.
\newblock \href
  {https://proceedings.neurips.cc/paper/2019/file/bdbca288fee7f92f2bfa9f7012727740-Paper.pdf}
  {Pytorch: An imperative style, high-performance deep learning library}.
\newblock In \emph{Advances in Neural Information Processing Systems},
  volume~32. Curran Associates, Inc.

\bibitem[{Pedregosa et~al.(2011)Pedregosa, Varoquaux, Gramfort, Michel,
  Thirion, Grisel, Blondel, Prettenhofer, Weiss, Dubourg, Vanderplas, Passos,
  Cournapeau, Brucher, Perrot, and {{\'E}}douard
  Duchesnay}]{JMLR:v12:pedregosa11a}
Fabian Pedregosa, Ga{{\"e}}l Varoquaux, Alexandre Gramfort, Vincent Michel,
  Bertrand Thirion, Olivier Grisel, Mathieu Blondel, Peter Prettenhofer, Ron
  Weiss, Vincent Dubourg, Jake Vanderplas, Alexandre Passos, David Cournapeau,
  Matthieu Brucher, Matthieu Perrot, and {{\'E}}douard Duchesnay. 2011.
\newblock \href {http://jmlr.org/papers/v12/pedregosa11a.html} {Scikit-learn:
  Machine learning in python}.
\newblock \emph{Journal of Machine Learning Research}, 12(85):2825--2830.

\bibitem[{Pennington et~al.(2014)Pennington, Socher, and
  Manning}]{Pennington2014GloVeGV}
Jeffrey Pennington, R.~Socher, and Christopher~D. Manning. 2014.
\newblock Glove: Global vectors for word representation.
\newblock In \emph{EMNLP}.

\bibitem[{Quionero-Candela et~al.(2009)Quionero-Candela, Sugiyama,
  Schwaighofer, and Lawrence}]{QuioneroCandela2009DatasetSI}
Joaquin Quionero-Candela, Masashi Sugiyama, Anton Schwaighofer, and Neil
  Lawrence. 2009.
\newblock Dataset shift in machine learning.

\bibitem[{Ramesh~Kashyap et~al.(2021)Ramesh~Kashyap, Mehnaz, Malik, Waheed,
  Hazarika, Kan, and Shah}]{ramesh-kashyap-etal-2021-analyzing}
Abhinav Ramesh~Kashyap, Laiba Mehnaz, Bhavitvya Malik, Abdul Waheed, Devamanyu
  Hazarika, Min-Yen Kan, and Rajiv~Ratn Shah. 2021.
\newblock \href {https://aclanthology.org/2021.adaptnlp-1.23} {Analyzing the
  domain robustness of pretrained language models, layer by layer}.
\newblock In \emph{Proceedings of the Second Workshop on Domain Adaptation for
  NLP}, pages 222--244, Kyiv, Ukraine. Association for Computational
  Linguistics.

\bibitem[{Sammut and Webb(2010)}]{ref1}
Claude Sammut and Geoffrey~I. Webb, editors. 2010.
\newblock \href {https://doi.org/10.1007/978-0-387-30164-8_832}
  {\emph{TF--IDF}}, pages 986--987. Springer US, Boston, MA.

\bibitem[{Sangodiah et~al.(2017)Sangodiah, Ahmad, and
  Ahmad}]{Sangodiah2017TAXONOMYBF}
A.~Sangodiah, Rohiza Ahmad, and W.~Ahmad. 2017.
\newblock Taxonomy based features in question classification using support
  vector machine 1.

\bibitem[{Sanh et~al.(2019)Sanh, Debut, Chaumond, and
  Wolf}]{Sanh2019DistilBERTAD}
Victor Sanh, Lysandre Debut, Julien Chaumond, and Thomas Wolf. 2019.
\newblock Distilbert, a distilled version of bert: smaller, faster, cheaper and
  lighter.
\newblock \emph{ArXiv}, abs/1910.01108.

\bibitem[{Sari et~al.(2020)Sari, Rini, and Malik}]{Sari2020TextCU}
Winda~Kurnia Sari, Dian~Palupi Rini, and R.~F. Malik. 2020.
\newblock Text classification using long short-term memory with glove features.

\bibitem[{Schwenk et~al.(2017)Schwenk, Barrault, Conneau, and
  LeCun}]{Schwenk2017VeryDC}
Holger Schwenk, Lo{\"i}c Barrault, Alexis Conneau, and Y.~LeCun. 2017.
\newblock Very deep convolutional networks for text classification.
\newblock In \emph{EACL}.

\bibitem[{Swart and Daneti(2019)}]{Swart2019AnalyzingLO}
A.~Swart and M.~Daneti. 2019.
\newblock Analyzing learning outcomes for electronic fundamentals using
  bloom’s taxonomy.
\newblock \emph{2019 IEEE Global Engineering Education Conference (EDUCON)},
  pages 39--44.

\bibitem[{Vaswani et~al.(2017)Vaswani, Shazeer, Parmar, Uszkoreit, Jones,
  Gomez, Kaiser, and Polosukhin}]{Vaswani2017AttentionIA}
Ashish Vaswani, Noam~M. Shazeer, Niki Parmar, Jakob Uszkoreit, Llion Jones,
  Aidan~N. Gomez, Lukasz Kaiser, and Illia Polosukhin. 2017.
\newblock Attention is all you need.
\newblock \emph{ArXiv}, abs/1706.03762.

\bibitem[{Wang et~al.(2018)Wang, Singh, Michael, Hill, Levy, and
  Bowman}]{Wang2018GLUEAM}
Alex Wang, Amanpreet Singh, Julian Michael, Felix Hill, Omer Levy, and
  Samuel~R. Bowman. 2018.
\newblock Glue: A multi-task benchmark and analysis platform for natural
  language understanding.
\newblock In \emph{BlackboxNLP@EMNLP}.

\bibitem[{Wolf et~al.(2020)Wolf, Debut, Sanh, Chaumond, Delangue, Moi, Cistac,
  Rault, Louf, Funtowicz, Davison, Shleifer, von Platen, Ma, Jernite, Plu, Xu,
  Le~Scao, Gugger, Drame, Lhoest, and Rush}]{wolf-etal-2020-transformers}
Thomas Wolf, Lysandre Debut, Victor Sanh, Julien Chaumond, Clement Delangue,
  Anthony Moi, Pierric Cistac, Tim Rault, Remi Louf, Morgan Funtowicz, Joe
  Davison, Sam Shleifer, Patrick von Platen, Clara Ma, Yacine Jernite, Julien
  Plu, Canwen Xu, Teven Le~Scao, Sylvain Gugger, Mariama Drame, Quentin Lhoest,
  and Alexander Rush. 2020.
\newblock \href {https://doi.org/10.18653/v1/2020.emnlp-demos.6} {Transformers:
  State-of-the-art natural language processing}.
\newblock In \emph{Proceedings of the 2020 Conference on Empirical Methods in
  Natural Language Processing: System Demonstrations}, pages 38--45, Online.
  Association for Computational Linguistics.

\bibitem[{Xue and Li(2015)}]{Xue2015ResearchOT}
Dashen Xue and Fengxin Li. 2015.
\newblock Research of text categorization model based on random forests.
\newblock \emph{2015 IEEE International Conference on Computational
  Intelligence \& Communication Technology}, pages 173--176.

\bibitem[{Yahya(2019)}]{Yahya2019SwarmIA}
Anwar~Ali Yahya. 2019.
\newblock Swarm intelligence-based approach for educational data
  classification.
\newblock \emph{J. King Saud Univ. Comput. Inf. Sci.}, 31:35--51.

\bibitem[{Yahya et~al.(2012)Yahya, Toukal, and Osman}]{Yahya2012BloomsTC}
Anwar~Ali Yahya, Z.~Toukal, and A.~Osman. 2012.
\newblock Bloom's taxonomy-based classification for item bank questions using
  support vector machines.
\newblock In \emph{Modern Advances in Intelligent Systems and Tools}.

\bibitem[{Yang et~al.(2016{\natexlab{a}})Yang, Yang, Dyer, He, Smola, and
  Hovy}]{Yang2016HierarchicalAN}
Zichao Yang, Diyi Yang, Chris Dyer, X.~He, Alex Smola, and E.~Hovy.
  2016{\natexlab{a}}.
\newblock Hierarchical attention networks for document classification.
\newblock In \emph{HLT-NAACL}.

\bibitem[{Yang et~al.(2016{\natexlab{b}})Yang, Yang, Dyer, He, Smola, and
  Hovy}]{yang-etal-2016-hierarchical}
Zichao Yang, Diyi Yang, Chris Dyer, Xiaodong He, Alex Smola, and Eduard Hovy.
  2016{\natexlab{b}}.
\newblock \href {https://doi.org/10.18653/v1/N16-1174} {Hierarchical attention
  networks for document classification}.
\newblock In \emph{Proceedings of the 2016 Conference of the North {A}merican
  Chapter of the Association for Computational Linguistics: Human Language
  Technologies}, pages 1480--1489, San Diego, California. Association for
  Computational Linguistics.

\bibitem[{Zorluoğlu et~al.(2019)Zorluoğlu, Bağrıyanık, and
  Şahintürk}]{Zorluolu2019AnalyzeOT}
S.~L. Zorluoğlu, Kübra~Elif Bağrıyanık, and Ayşe Şahintürk. 2019.
\newblock Analyze of the science and technology course teog questions based on
  the revised bloom taxonomy and their relation between the learning outcomes
  of the curriculum.
\newblock \emph{The International Journal of Progressive Education},
  15:104--117.

\end{thebibliography}


\end{document}